\documentclass[journal,twoside,web]{ieeecolor}
\usepackage{tmi}
\usepackage{cite}
\usepackage{amsmath,amssymb,amsfonts}
\usepackage{graphicx}
\usepackage{textcomp}
\usepackage{color}

\usepackage{caption}

\usepackage{tabu}                     % 表格插入
\usepackage{multirow}                 % 一般用以设计表格，将所行合并
\usepackage{multicol}                 % 合并多列
\usepackage{multirow}                 % 合并多行
\usepackage{float}                    % 图片浮动
\usepackage{makecell}                 % 三线表-竖线
\usepackage{booktabs}                 % 三线表-短细横线

\usepackage{amssymb}				  %	对勾
\usepackage{bbding}
\usepackage{pifont}
\usepackage{wasysym}
\usepackage{utfsym}

\usepackage{algorithm}				  %伪代码包  
\usepackage{algorithmicx}  
\usepackage{algpseudocode}

\usepackage{amsfonts,amssymb}		  %数学公式核心包
\usepackage{amsmath}

\usepackage{subcaption}               %子图
\usepackage{subfloat}

\usepackage{flushend}				  %最后一页末尾对齐

\def\BibTeX{{\rm B\kern-.05em{\sc i\kern-.025em b}\kern-.08em
		T\kern-.1667em\lower.7ex\hbox{E}\kern-.125emX}}
\begin{document}
	\title{Cross-Organ Domain Adaptive Neural Network for Pancreatic Endoscopic Ultrasound Image Segmentation}
	\author{ZhiChao Yan, Hui Xue, Yi Zhu, Bin Xiao, Hao Yuan
		\thanks{This work was supported by the National Natural Science	Foundation of (No. 62476056, 62076062, and 62306070), the Social Development Science and Technology Project of Jiangsu Province (No. BE2022811) and the Postgraduate Research \& Practive Innovation Program of Jiangsu Province (Grant number: KYCX24\_0415). Furthermore, the work was also supported by the Big Data Computing Center of Southeast University. (\textit{Corresponding author:Hui Xue})}
		\thanks{Zhichao Yan and Hui Xue are with the School of Computer Science and Engineering, Southeast University, Nanjing, 210096, China, and also with Key Laboratory of New Generation Artificial Intelligence Technology and Its Interdisciplinary Applications (Southeast University) Ministry of Education, China. (e-mail: zhichao\_yan@seu.edu.cn, hxue@seu.edu.cn).}
		\thanks{Yi Zhu, Bin Xiao and Hao Yuan are with Pancreas Centre, First Affiliated Hospital, Nanjing Medical University, and also with Department of General Surgery, First Affiliated Hospital, Nanjing Medical University, Nanjing, 210096, China.(e-mail: zhuyijssry@njmu.edu.cn, xiaobin@jsph.org.cn, yuanhao@njmu.edu.cn)}}
	
	\maketitle
	
	\begin{abstract}
		%由于胰腺癌早期缺乏特异性临床表现，因此预后较差。准确分割内窥镜超声（EUS）图像中的胰腺病变对诊断和治疗至关重要。虽然深度学习，尤其是卷积神经网络（CNN），推动了医学图像分割的发展，但传统的 CNN 假定训练集（源域）和测试集（目标域）之间的数据分布相同。在实践中，由于医学成像角度和协议的不同而造成的域偏差，这一假设往往是无效的。域自适应（DA）作为一种解决方案已受到关注，但现有的域自适应方法通常只关注单一器官，忽略了不同器官肿瘤的同质性特征。
		%为了解决这些问题，我们提出了跨器官肿瘤分割网络（COTS-Net），通过利用其他器官肿瘤的图像数据来提高胰腺 EUS 病变的分割准确性。我们的模型包括一个辅助网络和一个通用网络，前者用于完善特定领域的知识并改进表征学习，后者则利用这些完善的知识进行特定领域的共享。我们在通用网络中引入了一致性损失和边界损失，以捕捉不同器官肿瘤数据集之间的同质性，解决它们的异质性问题。
		%此外，我们还开发了胰腺癌内窥镜超声（PCEUS）数据集，其中包括 501 张病理证实的胰腺标注 EUS 图像，用于模型开发。结果表明，COTS-Net 能显著提高胰腺癌诊断的准确性。
	
%		Pancreatic cancer has received widespread attention as the malignant tumor with the highest mortality rate. 
		Accurate segmentation of lesions in pancreatic endoscopic ultrasound (EUS) images is crucial for effective diagnosis and treatment.
		However, the collection of enough crisp EUS images for effective diagnosis is arduous.
		Recently, domain adaptation (DA) has been employed to address these challenges by leveraging related knowledge from other domains.
		Most DA methods only focus on multi-view representations of the same organ, which makes it still tough to clearly depict the tumor lesion area with limited semantic information.
		Although transferring homogeneous similarity from different organs could benefit the issue, there is a lack of relevant work due to the enormous domain gap between them.
		To address these challenges, we propose the Cross-Organ Tumor Segmentation Networks (COTS-Nets), consisting of a universal network and an auxiliary network.
		The universal network utilizes boundary loss to learn common boundary information of different tumors, enabling accurate delineation of tumors in EUS despite limited and low-quality data.
		Simultaneously, we incorporate consistency loss in the universal network to align the prediction of pancreatic EUS with tumor boundaries from other organs to mitigate the domain gap.
		%aux网络通过..从而帮助univer网络提取域不变的知识
		To further reduce the cross-organ domain gap, the auxiliary network integrates multi-scale features from different organs, aiding the universal network in acquiring domain-invariant knowledge.
		Systematic experiments demonstrate that COTS-Nets significantly improves the accuracy of pancreatic cancer diagnosis.
		Additionally, we developed the Pancreatic Cancer Endoscopic Ultrasound (PCEUS) dataset, comprising 501 pathologically confirmed pancreatic EUS images, to facilitate model development.

	\end{abstract}
	
	\begin{IEEEkeywords}
		Pancreatic cancer, endoscopic ultrasound, medical image segmentation, cross-organ.
	\end{IEEEkeywords}

	\begin{figure}[!th]
		\centering
		\includegraphics[width=8.0cm]{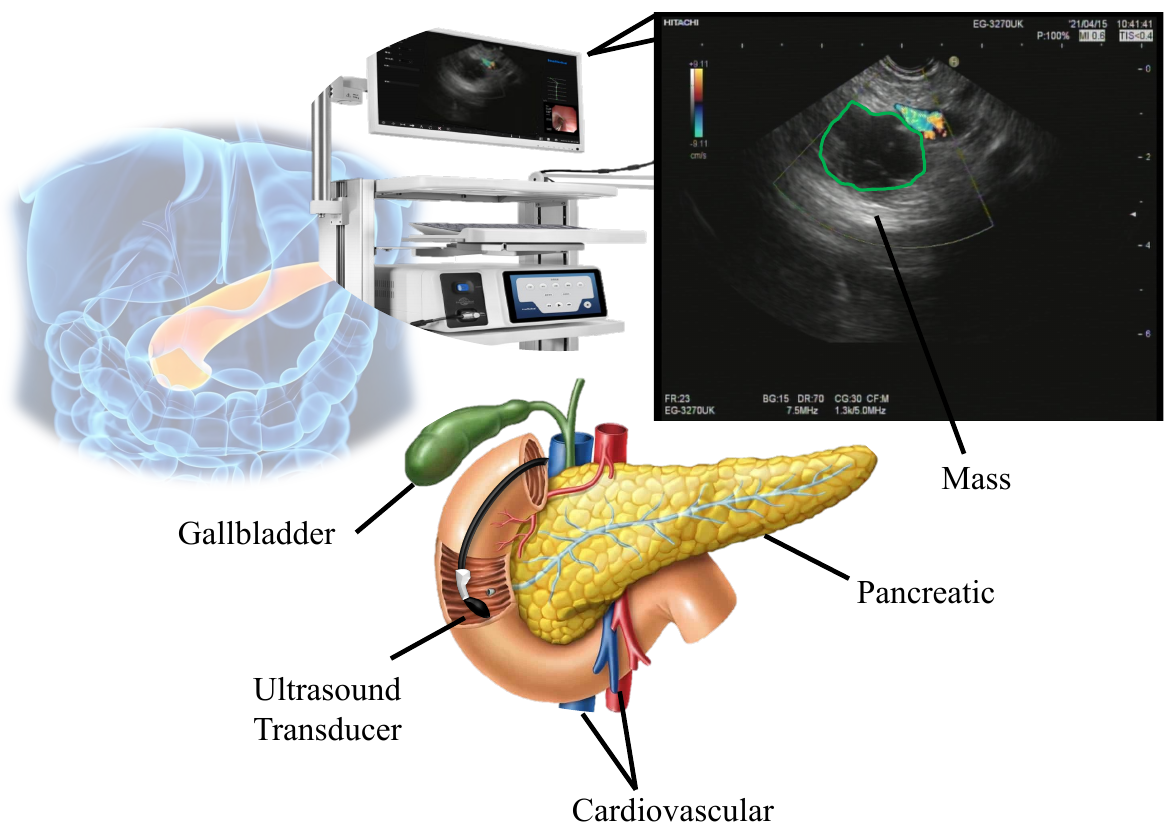}
		\caption{The endoscopic ultrasound (EUS) procedure for detecting pancreatic masses involves using the EUS device to examine the pancreatic lesion. The area depicted by the green curve in the figure is the identified mass.}
		\label{fig:Pancreatic_EUS}
	\end{figure}
	
	\section{Introduction}
	\label{sec:introduction}
	%总起胰腺癌预测的重要性
	\IEEEPARstart{P}{ancreatic} cancer is extremely malignant and lacks specific clinical manifestations in its early stages, resulting in a five-year survival rate of less than 7\% \cite{siegel2019cancer, wolrab2022lipidomic}. This is a serious problem for patients, and its early prediction and diagnosis are crucial for improving patient survival. To achieve this goal, computer-aided diagnosis (CAD) has been increasingly integrated into medical image analysis, demonstrating significant potential in pancreatic cancer detection \cite{barata2021explainable}. Traditional medical image segmentation techniques primarily rely on model-driven strategies, such as active contours, level sets, deformable models, and statistical shape models, which usually require extensive human intervention. The advent of deep learning has automated these processes, significantly enhancing diagnostic efficiency and accuracy \cite{noble2006ultrasound, lin2021seg4reg}.
	
	%现阶段针对胰腺癌的都是CT和MRI。随着技术的发展，现阶段EUS为较为主流的胰腺癌发现手段。EUS在
	
	%虽然这些成像技术能有效检测胰腺病变，但误诊率也很高。内镜超声（EUS）正成为胰腺癌诊断的重要工具（见图）。研究表明，在评估胰腺癌的局部浸润和区域淋巴结转移方面，EUS比CT更准确，尤其是在早期发现方面（cite{catalano2009eus}）。
	%对于疑似胰腺小肿瘤患者，EUS在检测这些肿瘤方面的灵敏度高达94.4%，明显高于对比增强CT 50%的灵敏度（cite{erickson2004eus}）。例如，在胰管扩张或狭窄扭曲的病例中，CT 图像可能无法显示病理，而超声图像则可以。
	%然而，获取胰腺 EUS 图像需要由专业医生进行精细操作，而且分辨率通常低于 CT 和 MRI 图像，边界和内部结构不明显，因此对胰腺癌医学图像进行分割具有挑战性。
	
	Currently, computed tomography (CT) and magnetic resonance imaging (MRI) are easily accessible, leading most pancreatic cancer studies to focus on training models using these modalities \cite{zhou2023meta, cao2023large}. 
%	Zhou \textit{et al}.  proposed a meta-information-aware dual-path Transformer for classifying and segmenting CT images of pancreatic lesions.
%	Cao \textit{et al}. \cite{cao2023large} developed a deep learning-based contrast-free CT pancreatic cancer detection (PANDA) method for pancreatic ductal adenocarcinoma (PDAC) detection and classification.
	However, due to the fixed and inflexible CT and MRI imaging views, the rate of misdiagnosis is also high. Endoscopic ultrasound (EUS) is emerging as a crucial tool for pancreatic cancer diagnosis (see Fig. \ref{fig:Pancreatic_EUS}). 
%	Research indicates that EUS is more accurate than CT in assessing local invasion and regional lymph node metastasis of pancreatic cancer, particularly for early detection \cite{catalano2009eus}. 
%	For patients with suspected small pancreatic tumors, EUS has demonstrated a sensitivity of 94.4\% in detecting these tumors, significantly higher than the 50\% sensitivity of contrast-enhanced CT \cite{erickson2004eus}. For example, in cases where there is pancreatic duct dilatation or narrowing and twisting, CT images may fail to reveal the pathology, while ultrasound images can.
%	However, acquiring EUS images of the pancreas requires delicate procedures performed by specialized physicians, and the resolution is usually lower than that of CT and MRI images, where borders and internal structures are not obvious, making it challenging to segment medical images for pancreatic cancer.
	Research indicates that EUS is accurate in assessing local invasion and regional lymph node metastasis of pancreatic cancer, particularly for early detection \cite{catalano2009eus}. 
	In patients with suspected small pancreatic tumors, EUS demonstrates a sensitivity of 94.4\% \cite{erickson2004eus}. Especially, in cases of dilated or narrowed and twisted pancreatic ducts, EUS images can reveal distinct lesion features.
%	However, acquiring pancreatic EUS images requires precise manipulation by specialized physicians, making the labeling of such data both expensive and challenging. Additionally, ultrasound images have low resolution and the boundaries and internal structures are not obvious, making segmentation of pancreatic cancer medical images challenging. Additionally, the low resolution of ultrasound images and the unclear boundaries and internal structures make the segmentation of pancreatic cancer particularly difficult. 
	However, EUS shows significant promise, limited research has been conducted due to the scare and low-quality data. 
	1) Data scarcity: Acquiring pancreatic EUS images requires precise manipulation by specialized physicians. 
	2) Low-quality: The boundaries between tumors and surrounding normal tissues are often indistinct.
	Therefore, it makes segmentation of EUS images particularly challenging.

	\begin{figure}[!t]
		\centering
		\includegraphics[width=8.0cm]{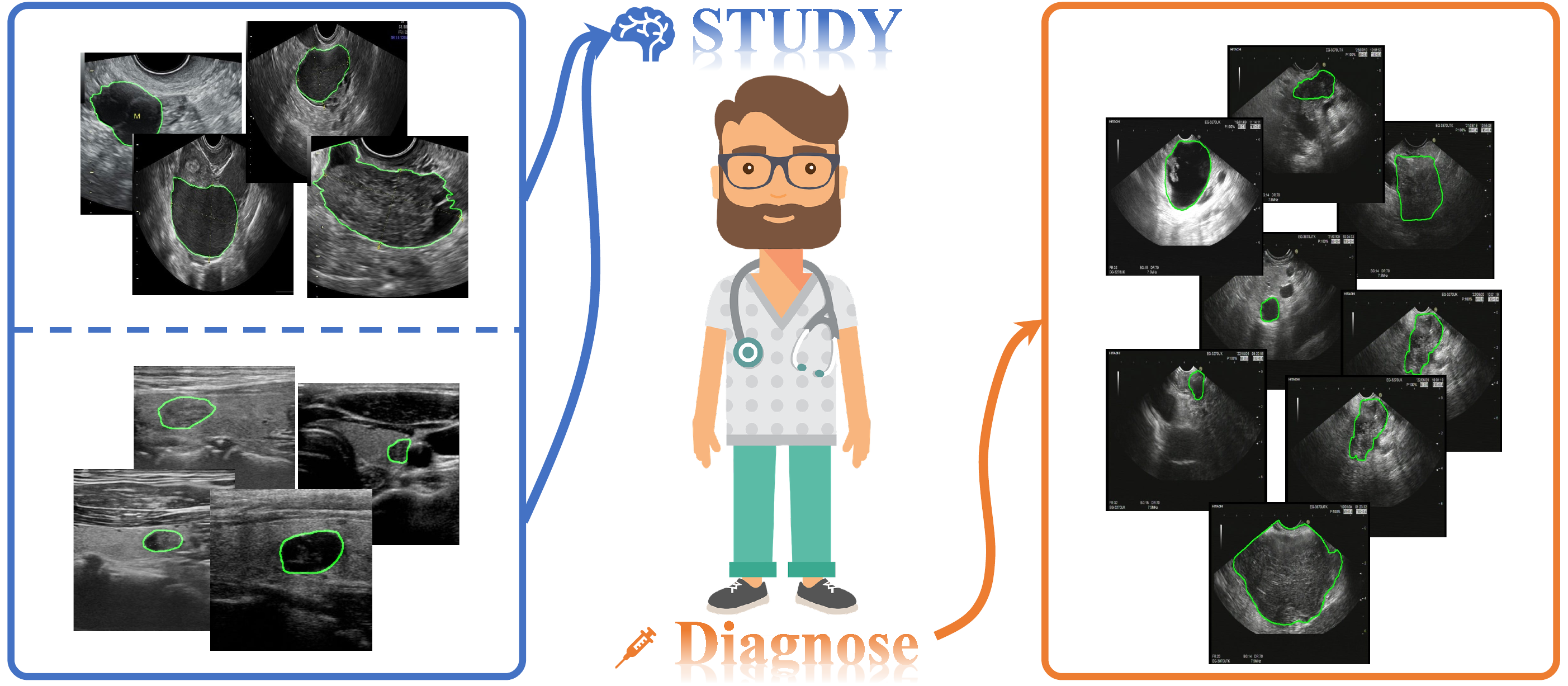}
		\caption{The training process for radiologists is not limited to a specific organ. Initially, they gain meta knowledge by learning to identify diagnostic lesions across various regions. After completing their studying, radiologists typically focus on diagnosing lesions in a specific organ.}
		\label{fig:Doctor_learning}
	\end{figure}

	%过渡到Domain adaptation的定义，提出现有研究大多数针对同一器官。然而经过研究表明，癌症细胞通过调控细胞-细胞连接和基质降解酶的表达，能够侵袭周围组织并形成转移。这种机制影响细胞在显微镜下的形态，表现为侵袭性细胞的形态特征的一致性。
	%重点突出我们的数据集是具有弱相关性，而不是同一个病种的强相关性
	Domain adaptation (DA) has shown great potential to solve the data scarcity problem in medical image analysis, leveraging knowledge from one domain (source domain) to improve performance in a different but related domain (target domain) \cite{pooch2020can, karani2018lifelong, albadawy2018deep, torralba2011unbiased, donahue2014decaf, ben2006analysis}. 
	By applying DA techniques, methods have been proposed for application in the medical field.
	Li \textit{et al}. \cite{GCNforPC} proposed an unsupervised domain adaptation segmentation framework based on GCN and meta-learning strategies, specifically for the task of segmenting a single organ—multiparametric pancreatic cancer MRI.
	Zhu \textit{et al}. \cite{zhu2019boundary} proposed a boundary-weighted domain adaptive network (BOWDA-Net) for automatic segmentation of prostate MR Images. This network is specifically designed to address the challenges of unclear boundaries and insufficient data in single-organ segmentation by using boundary-weighted segmentation loss and transfer learning.
	However, existing DA methods fail to effectively address EUS data due to the concentrating on multi-view or multi-modal representations of the same organ, resulting in limited feature extraction.

	Recently, some studies have demonstrated that tumors across different organs exhibit microscopic morphological consistency through the regulation of intercellular junctions and expression of matrix-degrading enzymes \cite{hanahan2011hallmarks}. 
	As shown in Fig. \ref{fig:Doctor_learning}, ultrasound images of different organs can improve a physician's ability to identify a specific organ by learning the homogeneous features of a tumor from different organs. 
	However, utilizing this knowledge from different organs in CAD is challenging due to the presence of a domain gap.
	Consequently, how to effectively use additional cross-organ tumor data for training to improve the segmentation performance on specific target tumors remains an urgent issue to be addressed.
	
	Motivated by the above-mentioned challenges, we propose a cross-organ tumor segmentation network (COTS-Nets) that mimics the diagnostic learning process of radiologists. 
	Our method comprises a universal network and an auxiliary network. 
	In the universal network, we introduce boundary loss. This loss enables the model to learn more accurate representations by emphasizing boundary features. As a result, the accuracy of pancreatic cancer EUS segmentation is improved.
	We also implement a consistency loss-based exponential moving average (EMA). This loss function guides the model to align the predictions of pancreatic EUS with tumor boundaries from other organs, alleviating the domain gap.
	%需要修改，辅助网络通过..从而帮助univer网络提取域不变的知识
	Additionally, to address the cross-organ domain gap, the auxiliary network integrates multi-scale features by extracting homogeneous features from different organs. These features subsequently help the universal network gain domain-invariant knowledge, achieving accurate segmentation in low-quality pancreatic EUS images.
	The main contributions of our method are:
	\begin{itemize}
%		\item[$\bullet$] To the best of our knowledge, we are the first to extract the homogeneous feature between the tumors of pancreatic and other organs. Meanwhile, our paradigm mimics the professional training process of radiologists by learning inherent diagnostic features such as echogenicity, tumor boundary, and texture from various types of images.
		\item[$\bullet$] We propose that the cross-organ paradigm mimics the professional training of radiologists, enabling them to learn inherent diagnostic features—such as echoes, tumor boundaries, and texture—from diverse image types. To the best of our knowledge, we are the first to extract the homogeneous features of pancreatic tumors with those from other organs.
		\item[$\bullet$] We propose a novel paradigm COTS-Nets. The universal network utilises tumour boundary correlation information to accurately delineate EUS lesion regions in data-limited scenarios. Subsequently, the auxiliary network enhances representation learning by fusing multi-scale features to refine domain-invariant knowledge in the auxiliary network.
		\item[$\bullet$] Furthermore, we propose a new pancreatic cancer endoscopic ultrasound image segmentation dataset “PCEUS”, which consists of 501 EUS detection images of pancreatic with pathologically confirmed labels.
	\end{itemize}
	%LEPset
	
	The remainder of this paper is organized as follows: In Section II, we introduce the works related to this study. In Sections III and IV, we describe our proposed methodology and the experimental results. Finally, Section V presents the conclusions.
	
	\begin{figure*}[ht]
		\centering
		\includegraphics[width=17.5cm]{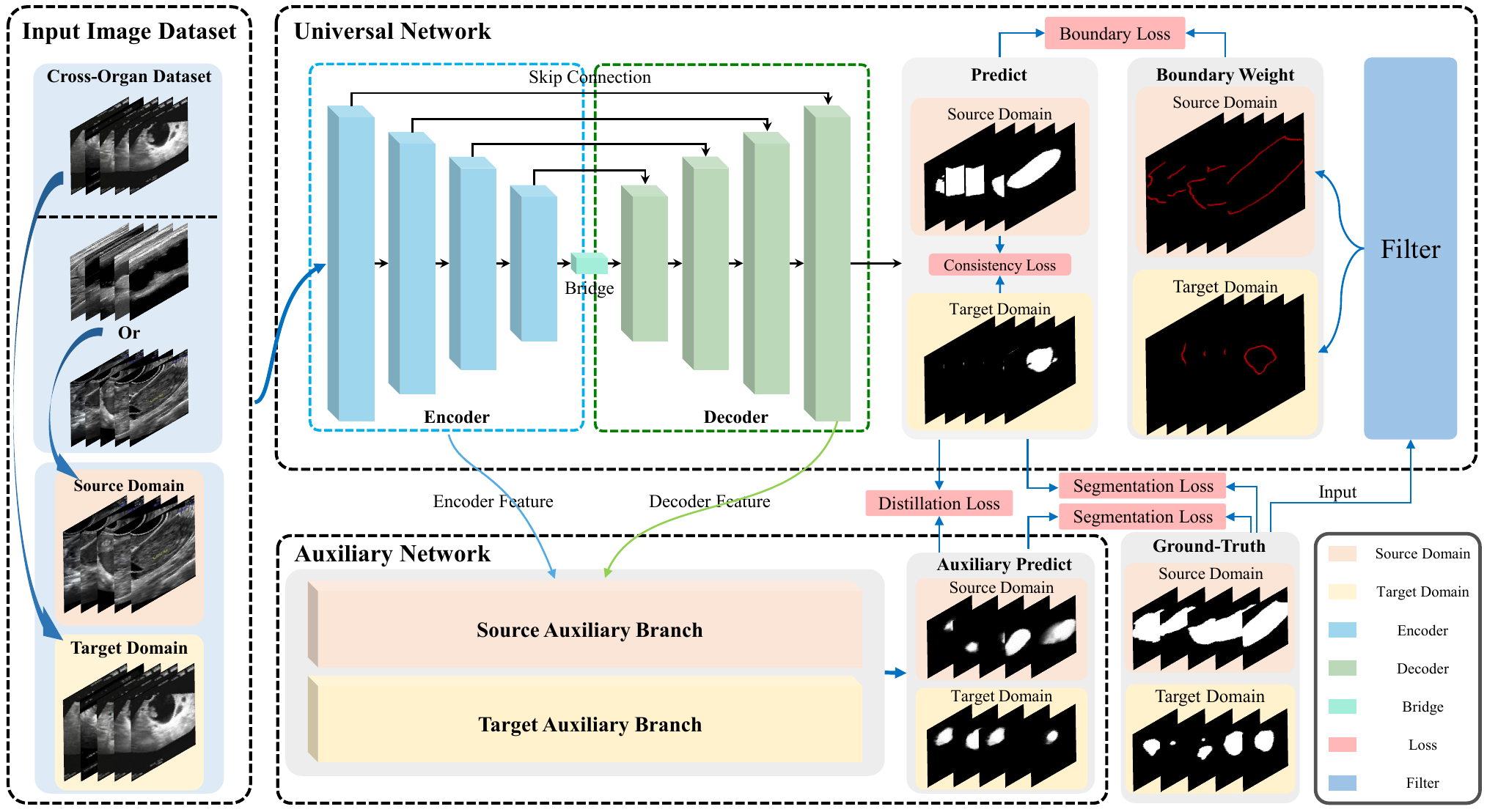}
		\caption{The schematic illustration of our proposed Cross-Organ Tumor Segmentation Networks (COTS-Nets), which is trained on cross-organ data using encoder-decoder with auxiliary network. The universal network comprises both encoder and decoder components to mitigate cross-organ heterogeneity. In each training iteration, the framework processes multiple batches of images, with each batch corresponding to a different domain. The universal network is trained with the supervision of ground-truth labels and the transferred knowledge from the auxiliary branches. }
		\label{fig:Method}
	\end{figure*}

	%%%%%%%%%%%%%%%%%%%%%%%%%%%%%%%%%%%%%%%%%%%%%%%%%%%%%%%%%%%%%%%%%%%%%%%%%%%%%%%%%%%%%%%%%%%%%%%%%%%%%%%%%%%%%%%%%%%%%%%%%%%%%%%%%%%%%%%%%%%%%%%%%%%%%%%%%%%%%%%%%%%%%%%%%%%%%%%%%%%%%%%%%%%%%%%%%%%%%%%%%%%%%%%%%%%%%%%%%%%%%%%%%%%%%%%%%%%%%%%%%%%%%%
	\section{Related Work}
	\subsection{Pancreatic Cancer Image Segmentation}
	Accurate segmentation of tumors from pancreatic images is crucial in medical imaging for preoperative diagnosis, surgical planning, and postoperative monitoring. However, due to the complex anatomy of the pancreas. Conventional methods try to deal with these problems using deep learning. For example, in order to fully mine representative semantic information from original CT images,  
	Cao \textit{et al}. \cite{cao2024strongly} proposed the strongly representative semantic guided segmentation network (SRSNet), which utilizes the intermediate semantic information to generate the highly representative high-resolution pre-segmented images, effectively reduces the channel redundancy between different resolutions.
	Li \textit{et al}. \cite{li2024deep}	employed a causality-aware module and counterfactual loss to enhance the deep learning network's understanding of the anatomical causality between foreground elements (pancreas and tumor) and background for synergistic segmentation of pancreas and tumor in 3D CT sequences.
	Qu \textit{et al}. \cite{qu2023transformer} proposed a tapered fusion network (TGPFN) that utilizes the global representations captured by the transformer to complement the long-distance-dependent transformations lost by convolutional operations at different resolutions.
	
	Although there are many studies of the above CT and MRI images, diagnostic accuracy remains low due to the fixed and inflexible imaging perspectives.
	EUS provides more accurately assessment of local infiltration in pancreatic cancer. However, relevant studies are scarce due to limited data availability.
	
	\subsection{EUS Computer-aided Diagnosis}
	EUS has a low misdiagnosis rate when detecting pancreatic tumors, making it particularly valuable for early diagnosis. However, acquiring EUS images requires high operator skill, and the images often suffer from artifacts, low contrast, and poor spatial resolution. Despite these challenges, many researchers have experimented with developing learning algorithms for EUS images.
	Kuwahara \textit{et al}. \cite{kuwahara2019usefulness} exploring whether artificial intelligence via deep learning algorithms using EUS images of intraductal papillary mucinous neoplasms (IPMNs) can predict malignancy.
	Zhang \textit{et al}. \cite{zhang2020deep} constructed an artificial intelligence system (pancreaticobiliary master) based on deep learning for pancreatic endoscopy with six classifications, where each classification corresponds to a location.
	Ozkan \textit{et al}. \cite{ozkan2016age} classified pancreatic and non-pancreatic cancers by segmenting pancreatic endoscopic ultrasound images into regions of interest (ROIs) using Relief-F to reduce the 122 extracted features, and inputting the 20 most favorable features identified into an artificial neural network for pancreatic and non-pancreatic cancer classification.
	
	While tumor classification is achievable, the amount of data is sparse and edge delineation is difficult. Consequently, limited work has been conducted on image segmentation, and the segmentation task is particularly challenging.
	
	\subsection{Domain Adaptation in Medical Image Segmentation}
	While convolutional neural networks (CNNs) have been successful in medical image segmentation, their ability to generalize may be limited due to the vast differences in medical imaging data across scanners, protocols, and patient populations. 
	
	Recently, researchers have applied domain adaptation using deep neural networks to medical image analysis. This approach aims to enhance model generalization to new or unseen datasets, improving robustness and accuracy across different clinical settings. 
	Liu \textit{et al}. \cite{liu2020shape} introduced a shape-aware meta-learning scheme that optimizes model performance by simulating domain shifts during training, which uses objectives—shape compactness and shape smoothing loss to improve the generalization of prostate MRI segmentation models.
	Basak \textit{et al}. \cite{basak2023semi} proposed a semi-supervised domain adaptive (SSDA) approach for medical image segmentation. Their method involves a two-stage training process: first, the encoder undergoes pre-training via contrast learning with domain content decoupling, followed by fine-tuning of the encoder and decoder for pixel-level segmentation in a semi-supervised setting.
	Yu \textit{et al}. \cite{yu2023source}
	proposed a shape-aware meta-learning scheme that optimizes model performance by simulating domain shifts during training, which uses shape compactness loss and shape smoothing loss to improve the generalization of prostate MRI segmentation models.
	
	However, existing domain adaptation methods predominantly address domain shifts arising from variations in imaging modalities or acquisition protocols for the same anatomical region. They often overlook the distinct characteristics and segmentation challenges presented by different organs, limiting their ability to generalize effectively across diverse anatomical targets.
	
	\section{Method}
	\subsection{Preliminary}
	%	Formally, we denote a labeled source domain dataset $\mathcal{D}_{i}^{s}=\{x_{i}^{s},y_{i}^{s}\}_{i}^{N_{s}}$, where $x_{i}^{s}$ is the $i$-th image from the source domain, $y_{i}^{s}$ is the corresponding pixel-wise annotation of $x_{i}^{s}$, and $N_{s}$ is the total number of source domain samples. In addition, we also have access to the target domain $\mathcal{D}_{i}^{t}=\{x_{i}^{t},y_{i}^{t}\}_{i}^{N_{t}}$. We aim to train a segmentation model $S_{\theta}: x\rightarrow y$ to make the model perform well in the target data $x_{i}^{t}\in \mathcal{D}_{i}^{t}$.
	Formally, we denote a labeled source domain dataset as $\mathcal{D}^{s}=\{x_{i}^{s},y_{i}^{s}\}_{i}^{N_{s}}$, where $x_{i}^{s}$ is the $i$-th image from the source domain, $y_{i}^{s}$ is the corresponding pixel-wise annotation of $x_{i}^{s}$, and $N_{s}$ is the total number of source domain samples. The source domain images primarily come from one or more specific organs, and these images have been annotated by experts, ensuring high-quality pixel-wise annotations.
	Additionally, we have access to a target domain dataset $\mathcal{D}^{t}=\{x_{i}^{t},y_{i}^{t}\}_{i}^{N_{t}}$, where $x_{i}^{t}$ is the $i$-th image from the target domain, $y_{i}^{t}$ is the corresponding pixel-wise annotation of $x_{i}^{t}$, and $N_{t}$ is the total number of target domain samples. Unlike the source domain, the target domain data come from different organs, implying that the distribution of these images and annotations might significantly differ from the source domain.
	The objective of this paper is to train a segmentation model $S_{\theta}: x \rightarrow y$ using source and partial target domain datasets, and then transfer this model to an unseen target domain. The goal is for the model to perform well on unseen target domain data $x_{i}^{t} \in \mathcal{D}^{t}$. Specifically, we aim for the model $S_{\theta}$ to accurately segment lesion areas in the target domain images, even though these images originate from organs different from those in the source domain.

	\subsection{Overview}
%	The overview of the COTS-Net framework is depicted in Fig. \ref{fig:Method}. Our COTS-Net consists of a universal network and an auxiliary network, which work in tandem to enhance segmentation performance across different organs. The universal network employs a combination of encoder-decoder architecture to effectively process and segment the input images. The auxiliary network, comprising source and target auxiliary branches, supplements the universal network by providing domain-invariant features and facilitating knowledge transfer between domains.
%	The synergistic guidance from the ground truth labels and the transferred cross-organ knowledge from the auxiliary branches enable the universal network to learn more robust and generalized representations. This shared homogeneity across organs allows the network to better adapt to variations in the target domain, ultimately improving segmentation accuracy and consistency.
	The overview of the COTS-Nets framework is depicted in Fig. \ref{fig:Method}. 
	Our COTS-Nets consists of a universal network and an auxiliary network, which work in tandem to enhance segmentation performance across different organs. The universal network employs a combination of encoder-decoder architecture to effectively process and segment the input images. The auxiliary network, comprising source and target auxiliary branches, supplements the universal network by providing domain-invariant features and facilitating knowledge transfer between domains.
	The synergistic guidance from the ground-truth labels and the transferred cross-organ knowledge from the auxiliary branches enable the universal network to learn more robust and generalized representations. This shared homogeneity across organs allows the network to better adapt to variations in the target domain, ultimately improving segmentation accuracy and consistency.

	\subsection{Universal Network for Robust Segmentation}
	The universal network forms the core segmentation architecture of COTS-Nets. It is designed to handle the diverse and complex features present in medical images from different organs. By leveraging a robust encoder-decoder framework, the universal network processes input images and generates accurate segmentation masks. This network aims to extract homogeneous features across different tumors, ensuring reliable and consistent performance can be obtained for pancreatic cancer.
	The universal network incorporates updates for segmentation loss, boundary loss, consistency loss, and distillation loss, with the distillation loss being detailed in the Section \ref{section:A}.
	
	% Boundary Knowledge Transfer Loss
	To achieve significantly accurate segmentation for cross-organ tumors, we propose a boundary loss between the prediction of universal network and ground-truth. 
	Let $\{x_{i}^{s}, y_{i}^{s}\}\in \mathcal{D}^{s}$ represents the training images and ground-truth from the source domain, which $\{x_{i}^{t}, y_{i}^{t}\}\in \mathcal{D}^{t}$ can also access.
	To obtain the boundary map generated from the ground-truth, we use the Sobel filter and Gaussian filter. Define the Sobel filter as $S_{x}$ and $S_{y}$, the gradients $G^{(\cdot)}_{i}$ of the $i$-th ground-truth from the source or target domain is calculate as follow:
	\begin{equation}
		\begin{split}
			%			G_{i,x}^{s} = S_{x} * y_{i}^{s},\\
			%			G_{i,x}^{s} = S_{y} * y_{i}^{s},\\
			G^{(\cdot)}_{i} = \sqrt{{G_{i,x}^{(\cdot)}}^{2}+{G_{i,y}^{(\cdot)}}^{2}} 
		\end{split}
	\end{equation}
	where $G_{i,x}^{(\cdot)} = S_{x} * y_{i}^{(\cdot)}$ and $G_{i,x}^{(\cdot)} = S_{y} * y_{i}^{(\cdot)}$ are the gradients in the $x$ and $y$ directions, $*$ denotes the convolution operation. 
	After completing the process of detecting the edges of the lesion by calculating the gradient of the image, the image is smoothed using a Gaussian filter to reduce the effect of noise on edge detection. Thus, the boundary map $M^{(\cdot)}_{i}$ is defined as:
	\begin{equation}
		\begin{split}
			M^{(\cdot)}_{i} = \frac{(G^{(\cdot)}_{i} * G_{ker})-min(G^{(\cdot)}_{i} * G_{ker})}{max(G^{(\cdot)}_{i} * G_{ker})-min(G^{(\cdot)}_{i} * G_{ker})}
		\end{split}
	\end{equation}
	where $*$ denotes the convolution operation. $M^{(\cdot)}_{i}$ is the $i$-th boundary map of source or target domain. $G_{ker}$ is the Gaussian kernel.
	
	In order to deals with the alignment of boundary features, we introduce the boundary loss:
	\begin{equation}
		\begin{split}
			\mathcal{L}_{bound}= -\mathbb{E}_{x^{s}_{i}\in \mathcal{D}^{s}}[(1+\gamma M^{s}_{i})log(\hat{Y}_{i,uni}^{s})] \\
			-\mathbb{E}_{x^{t}_{i}\in \mathcal{D}^{t}}[(1+\gamma M^{t}_{i})log(\hat{Y}_{i,uni}^{t})]
		\end{split}
	\end{equation}
	where $\gamma$ is a weighted coefficient.
	
	% EMA Loss/ Domain Attention Consistency Loss
	Boundary-based loss is capable of finely inscribing segmentation boundaries, but it mainly designs local features. However, in cross-organ medical image segmentation, the alignment of overall features is particularly important.
	In order to deal with the alignment of overall features, we thus use EMA to get a more accurate estimation of the mean shadow weights of the entire population (within each domain).
	%	consis_loss = consistency_loss(EMA_source, EMA_target)
	Specifically, the EMA shadow weights are computed as:
	\begin{equation}
		\begin{split}
			m_{s}=\lambda m_{s}+(1-\lambda)\frac{1}{B}\sum_{i=1}^{B}\hat{Y}_{i,uni}^{s}
		\end{split}
	\end{equation}
	\begin{equation}
		\begin{split}
			m_{t}=\lambda m_{t}+(1-\lambda)\frac{1}{B}\sum_{i=1}^{B}\hat{Y}_{i,uni}^{t}
		\end{split}
	\end{equation}
	where $\lambda$ is the weight value. $\hat{Y}_{i,uni}^{s}$ and $\hat{Y}_{i,uni}^{t}$ denote the prediction of $i$-th image in the source domain or target domain from universal network respectively. $B$ is the number of samples over the mini-batch.
	The consistency loss is computed as the mean-square error (MSE) between $m_{s}$ and $m_{t}$ aiming at extracting the consistency of tumor boundary features. Formally, the consistency loss is defined as:
	\begin{equation}
		\begin{split}
			%			\mathcal{L}_{consist}= \frac{1}{B}\sum_{i=1}^{B}(\hat{Y}_{i,uni}^{s}-\hat{Y}_{i,uni}^{t})^{2}
			\mathcal{L}_{consist}= \frac{1}{B}\sum_{i=1}^{B}(m_{s}-	m_{t})^{2}
		\end{split}
	\end{equation}

	\subsection{Auxiliary Network for Cross-Organ Knowledge Transfer}
	\label{section:A}
\begin{figure}[!t]
		\centering
		\includegraphics[width=6.8cm]{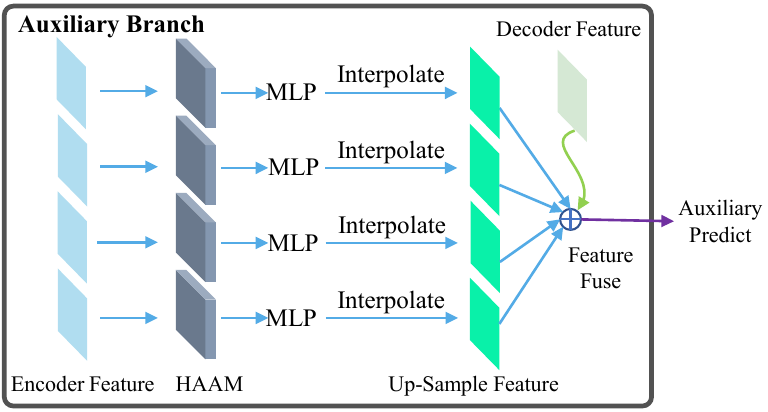}
		\caption{The components of auxiliary network branch.}
		\label{fig:Auxiliary_Branch}
	\end{figure}

	The auxiliary network enhances the capabilities of the universal network by incorporating domain-invariant knowledge. This network is composed of source and target auxiliary branches that provide additional features tailored to each domain. It is worth to note that the source and target auxiliary branch has the same architecture, as Fig. \ref{fig:Auxiliary_Branch}. 
	By transferring cross-organ knowledge and leveraging ground-truth labels, the auxiliary network ensures that the universal network learns more robust and generalized representations. This synergistic approach enables the model to adapt effectively to variations in target domain data.
	
	\begin{table*}[!t]
		\normalsize
		%		\large
		\centering
		\renewcommand\arraystretch{1.25}
		\caption{Comparison of the segmentation performance of COTS-Nets and other state-of-art approaches. $\uparrow$ means forward indicator, $\downarrow$ means reverse indicator.}
		\label{Table1}
		%		\resizebox{19cm}{!}
		\resizebox{\textwidth}{!}
		{
			\begin{tabular}{c|p{1.5cm}<{\centering}|p{1.5cm}<{\centering}|p{1.5cm}<{\centering}|p{1.5cm}<{\centering}|p{1.5cm}<{\centering}|p{1.5cm}<{\centering}|p{1.5cm}<{\centering}|p{1.5cm}<{\centering}}
				\toprule %添加表格头部粗线
				\multicolumn{1}{c|}{\multirow{2}{*}{Method}}&\multicolumn{4}{c|}{mmotu$\rightarrow$pancreatic}&\multicolumn{4}{c}{thyroid$\rightarrow$pancreatic}\\
				\cline{2-9}
				
				\multicolumn{1}{c|}{}& Dice $\uparrow$ & IoU $\uparrow$ & ASD $\downarrow$& HD95 $\downarrow$& Dice $\uparrow$ & IoU $\uparrow$ & ASD  $\downarrow$ & HD95 $\downarrow$\\
				
				\cline{1-9}
				\multicolumn{1}{c|}{PFNet}			& 78.326 & 68.249 & 5.45  & 18.03 & 76.346 & 66.300 & 6.58 & 19.18 \\
				\multicolumn{1}{c|}{FEDER}			& 78.836 & 68.097 & 6.23  & 16.43 & 78.829 & 68.429 & 5.43 & 16.14 \\
				
				\cline{1-9}
				
				\multicolumn{1}{c|}{U\_Net} 		& 78.943 & 68.264 & 0.86 & 9.98 & 78.716 & 68.738 & 1.75 &9.42 \\ 
				
				\multicolumn{1}{c|}{U\_Net v2} 		& 74.367 & 62.514 & 3.28 & 12.33 & 73.648 & 61.693 & 3.35 & 13.09 \\
				
				\multicolumn{1}{c|}{AAUNet} 		& 23.095 & 13.752 & 41.84 & 112.88 & 10.220 & 5.580 & 43.59 & 116.13 \\  
				
				\multicolumn{1}{c|}{Swin-Unet}		& 70.001 & 62.748 & 5.06 & 13.46 & 63.405 & 56.745 & 6.14 & 15.14 \\
				
				\multicolumn{1}{c|}{BA-Transformer} & 46.945 & 32.716 & 18.57 & 54.67 & 45.571 & 31.323 & 19.72 &  60.30   \\  
				
				\multicolumn{1}{c|}{ConvFormer} 	& 73.105 & 60.679  & 3.57  & 14.70 & 68.463 & 55.876 & 4.79 & 19.02 \\
				
				\multicolumn{1}{c|}{MDViT} 			& 79.881 & 67.007 & 0.62  & 4.94  & 79.862 & 67.035 & 0.83 & 6.06 \\

				\cline{1-9}
				\multicolumn{1}{c|}{\textbf{COTS-Nets \textit{(ours)}}}& \textbf{80.975} & \textbf{68.599} & \textbf{0.42}  & \textbf{4.76} & \textbf{81.652} & \textbf{69.504} & \textbf{0.71} &\textbf{4.07} \\
				\bottomrule %添加表格底部粗线
		\end{tabular}}
	\end{table*}

	% HAAM & MLP 
	% HAAM 通过channel dimension and spatial dimension约束从多个角度提取跨器官肿瘤的同质性。
	
	Considering the different organ tumors have specific domain representation, so inspired by the adaptive attention U-Net \cite{chen2022aau}, we use the hybrid adaptive attention module (HAAM) which can learn more robust representations from images through channel dimension and spatial dimension constraints to help extract the hybrid-specific feature. Meanwhile, we construct parallel networks in the auxiliary network to process the features extracted by encoders in different layers to obtain the inherent representations of lesions in the target domain and the source domain. 
	Given a medical image $x_{i}^{(\cdot)}$ from source or target domain, the encoder will extract the feature $X_{i, enc}^{l,(\cdot)}$ of $x_{i}^{(\cdot)}$ in $l$-th layers.
	The channel attention block output $H^{l,(\cdot)}_{i,chan}$ is defined as follow:

	\begin{equation}
		\begin{split}
			Act^{l,(\cdot)}_{i} = \mathrm{Sigmoid}(W_{2}\cdot \mathrm{ReLU}(W_{1}\cdot \mathrm{GAP}(X_{i,enc}^{l,(\cdot)})))
		\end{split}
	\end{equation}
	\begin{equation}
		\begin{split}
			H^{l,(\cdot)}_{chan} = Act^{l,(\cdot)}_{i} \cdot X_{i,enc}^{l,(\cdot)}
		\end{split}
	\end{equation}
	where $W_{1}, W_{2}$ are the learnable weight.
	And the spatial attention block will use the output of the channel attention block as input. Therefor, the final output $H^{l,(\cdot)}_{i,space}$ is:
	\begin{equation}
		\begin{split}
			H^{l,(\cdot)}_{i,space} = &\mathrm{Sigmoid}(\mathrm{Conv}(\mathrm{Concate}(\mathrm{AvgPool}(H^{l,(\cdot)}_{i,chan})), \\
			& \mathrm{MaxPool}(H^{l,(\cdot)}_{i,chan}))) \cdot H^{l,(\cdot)}_{i,chan}
		\end{split}
	\end{equation}
	
	% MLP的目的：将特征转换为非线性，提高原始特征的表达能力
	And then the output of HAAM is nonlinear transformed as input features using multilayer perception (MLP)\cite{xie2021segformer}, so as to enhance the expression ability of the original feature.  
	\begin{equation}
		\begin{split}
			F_{i,mlp}^{l,(\cdot)} = \mathrm{MLP}(H^{l,(\cdot)}_{i,space})
		\end{split}
	\end{equation}
	
	To unify the feature dimension $\frac{H}{4}\times \frac{W}{4}$ consistent with the original image dimension, interpolate is used to up-sample the $F_{i,mlp}^{l,(\cdot)}$. Then, the interpolated features $F_{i,interp}^{l,(\cdot)}=Up\-sampler(F_{i,mlp}^{l,(\cdot)})$ are concatenated in the channel dimension $F_{i,concat}^{(\cdot)} = \mathrm{Concate}(F_{i,interp}^{1,(\cdot)}, F_{i,interp}^{2,(\cdot)},...,F_{i,interp}^{L,(\cdot)})$, where $L$ is the number of encoder layer.
	To fuse features from different domains and branches together to enhance the final segmentation results, we concate $Y_{i,concat}^{(\cdot)}$ and $X_{i,dec}^{(\cdot)}$ from the last layer in decoder module. 
	Therefore, the auxiliary output is defined as:
	\begin{equation}
		\begin{split}
			\hat{Y}_{i,aux}^{(\cdot)} = F_{i,concat}^{(\cdot)} + X_{i,dec}^{(\cdot)}
		\end{split}
	\end{equation}
	where $\hat{Y}_{i,aux}^{(\cdot)}$ is the prediction of the auxiliary network from $i$-th image in source or target domain. 
	
	% Mutual Knowledge Distillation
	%	Distilling knowledge from domain-specific networks has been found beneficial for universal networks to learn more robust representations. Moreover, mutual learning that transfers knowledge between teachers and students enables both to be optimized simultaneously [15]. To realize these benefits, we propose MKD that mutually transfers knowledge between auxiliary peers and the universal network. In Fig. 1-d, the mth auxiliary peer is only trained on the mth domain, producing output ˆ Y m, whereas the universal network’s output is ˆ Y . 
	Similar to MS-Net \cite{liu2020ms}, we utilize the symmetric dice loss as a refinement loss.
	Each auxiliary branch acts as an expert in its specific domain and guides the universal network to learn domain-invariant information. The universal network can acquire homogenized features of cross-organ tumors from the expertise of each auxiliary branch, enabling cross-organ tumor knowledge sharing and preserving fine-grained details of the target domain data in the auxiliary branches.
	The distillation loss is:
	\begin{equation}
		\begin{split}
			\mathcal{L}_{distill} = \mathcal{L}_{Dice}^{u} (\hat{Y}_{i,aux}^{(\cdot)},\hat{Y}_{i,uni}^{(\cdot)})
		\end{split}
	\end{equation}
	where $\mathcal{L}_{distill}$ updates in the universal network.
	
	\subsection{Objective Function}
	The objective function for updating the universal network is calculated as follow:
	\begin{equation}
		\begin{split}
%			\mathcal{L}_{uni} = &\alpha \mathcal{L}_{seg}^{u} (Y,\hat{Y}_{i,uni}^{(\cdot)}) + \beta  \mathcal{L}_{distill} \\
%			&+ \mathcal{L}_{bound} + \mathcal{L}_{consist}
			\mathcal{L}_{uni} = &\alpha \mathcal{L}_{seg} (Y_{i}^{(\cdot)},\hat{Y}_{i,uni}^{(\cdot)})  + \beta  \mathcal{L}_{distill} \\
					& + \mathcal{L}_{bound} + \mathcal{L}_{consist}
		\end{split}
	\end{equation}
%	$\mathcal{L}_{seg}^{u} (Y_{i}^{(\cdot)},\hat{Y}_{i,uni}^{(\cdot)}) = \mathcal{L}_{Dice}^{(\cdot)}(Y_{i}^{(\cdot)},\hat{Y}_{i,uni}^{(\cdot)}) + \mathcal{L}_{bce}^{(\cdot)} (Y_{i}^{(\cdot)},\hat{Y}_{i,uni}^{(\cdot)})$

	The objective function for updating the auxiliary network is defined as:
	\begin{equation}
		\begin{split}
			%\mathcal{L}_{aux}=  \mathcal{L}_{seg}^{a} (Y,\hat{Y}_{i,aux}^{(\cdot)}) 
%			\mathcal{L}_{aux} = \mathcal{L}_{seg}^{a} (Y_{i}^{(\cdot)},\hat{Y}_{i,aux}^{(\cdot)})
%			\mathcal{L}_{aux}=  \mathcal{L}_{Dice}^{(\cdot)}(Y_{i}^{(\cdot)},\hat{Y}_{i,aux}^{(\cdot)})  + \mathcal{L}_{bce}^{(\cdot)} (Y_{i}^{(\cdot)},\hat{Y}_{i,aux}^{(\cdot)})
			\mathcal{L}_{aux} = \mathcal{L}_{seg} (Y_{i}^{(\cdot)},\hat{Y}_{i,aux}^{(\cdot)})
		\end{split}
	\end{equation}
	where $(\cdot)$ represents source or target domain. $\mathcal{L}_{seg} =\mathcal{L}_{Dice} + \mathcal{L}_{bce}$ combines Dice and binary cross	entropy loss. $\alpha$ and $\beta$ are the weight value. $Y_{i}^{(\cdot)}$ is the matrix of the $i$-th ground-truth image in the source or target domain.
	%这里的超参数需要确定下
	
	%	\begin{algorithm}[!ht]
		%		\caption{COTS-Net for Cross-Organ Tumor Segmentation} % 算法的名字
		%		\label{alg1}
		%		\hspace*{0.02in} {\bf Input:} % 算法的输入
		%		\\ \hspace*{0.02in} Training data $X$;
		%		\\ \hspace*{0.02in} Ground truth labels $Y$;
		%		\\ \hspace*{0.02in} Number of epochs $N$;
		%		\\ \hspace*{0.02in} Learning rate $\alpha$;
		%		\\ \hspace*{0.02in} Batch size $B$;
		%		\\\hspace*{0.02in} Initialize network parameters $\theta$;
		%		\begin{algorithmic}[1]
			%			\State Pre-processing stage;
			%			\For{each image $x_i \in X$}
			%				\State Normalize image $x_i$;
			%				\State Apply data augmentation to $x_i$;
			%			\EndFor
			%			
			%			\State Training stage:
			%			\For{epoch $t = 1$ to $N$}
			%				\For{each mini-batch $(X_{batch}, Y_{batch})$}
			%					\State Extract features $f_i$ from $X_{batch}$ using the feature extractor $f(\cdot)$;
			%					\State Compute segmentation masks $\hat{Y}_{batch}$ using COTS-Net;
			%					\State Calculate loss $L$ using ground truth $Y_{batch}$ and predictions $\hat{Y}_{batch}$;
			%					\State Update network parameters $\theta$ using gradient descent: $\theta \leftarrow \theta - \alpha \nabla_\theta L$;
			%				\EndFor
			%			\EndFor
			%			
			%			\State Postprocessing stage:
			%			\For{each predicted mask $\hat{y}_i$}
			%				\State Apply postprocessing techniques (e.g., CRF) to refine $\hat{y}_i$;
			%			\EndFor
			%			
			%			\State Output final segmentation masks $\hat{Y}$;
			%		\end{algorithmic}
		%	\end{algorithm}

	\section{Experiments}

	\subsection{Dataset}
	
	\subsubsection{Ovarian Tumor Ultrasound Image Segmentation Dataset}Multi-Modality Ovarian Tumor Ultrasound (MMOTU) image dataset includes two subsets with distinct modalities: 1,469 2D ultrasound images and 170 contrast-enhanced ultrasound (CEUS) images. Both subsets provide pixel-wise semantic annotations and global category annotations. In this paper, we just use the MMOTU\_2d \cite{zhao2022multi}.
	
	\subsubsection{Thyroid Tumor Ultrasound Image Segmentation Dataset}TN3K dataset is open-access and includes 3,493 grayscale ultrasound images from 2,421 patients. Each image has been cropped to remove non-ultrasound areas and is annotated with high-quality nodule masks from various devices and views. In this article, we refer to it as Thyroid\_tn3k \cite{Thyroid}.
	
	\subsubsection{Pancreatic Tumor Ultrasound Image Segmentation Dataset} To achieve effective segmentation of pancreatic tumors, we constructed a pancreatic EUS segmentation dataset containing 501 EUS images, where the mask is labeled by a professional radiologist.
	
\begin{figure}[!t]
		\centering
		\includegraphics[width=8.0cm]{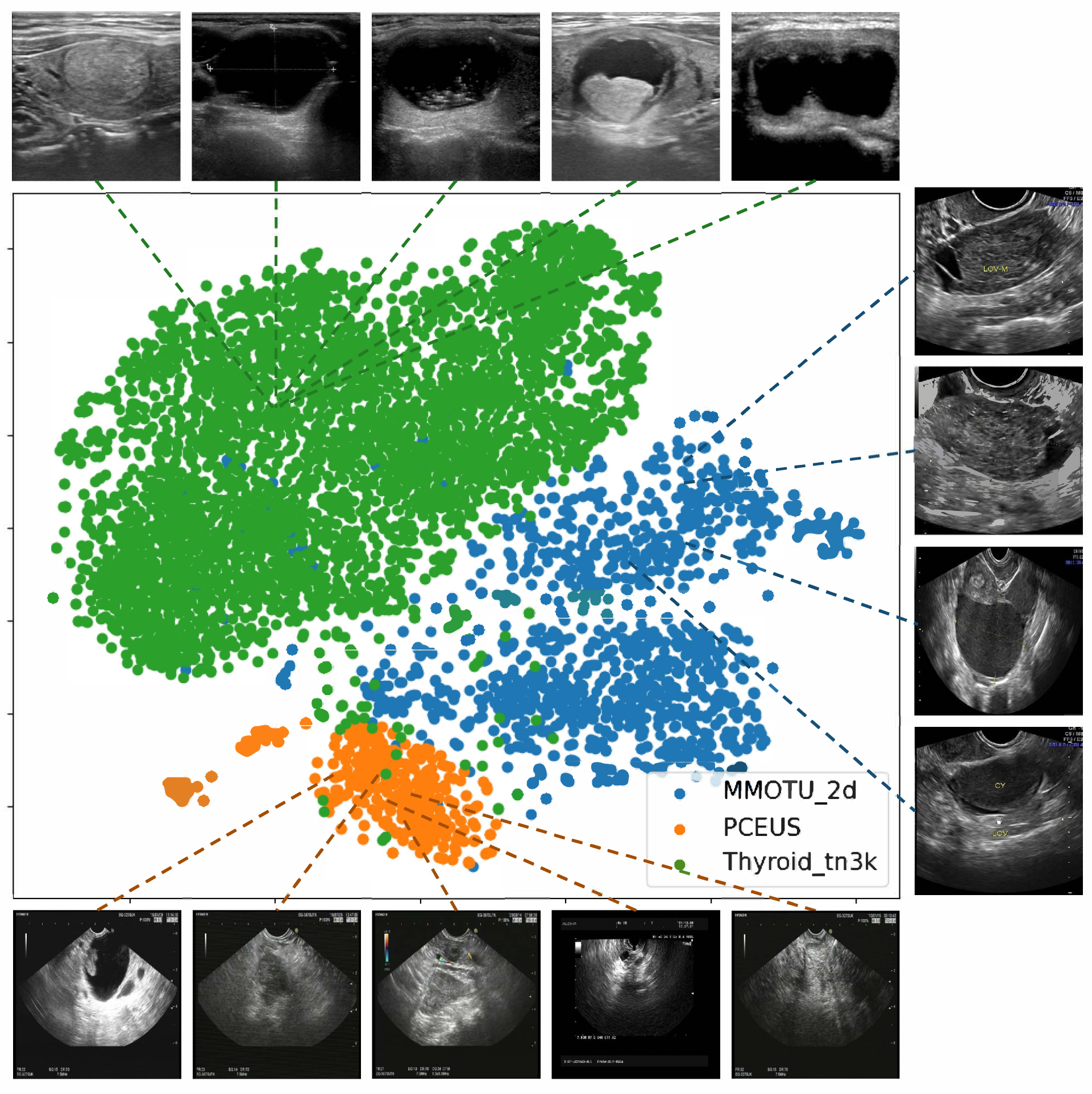}
		\caption{T-SNE distribution visualization. The MMOTU\_2d data is represented in green, the Thyroid\_tn3k data in blue, and our self-constructed PCEUS dataset in orange.}
		\label{fig:t-SNE}
	\end{figure}

	To visualize the distribution of the datasets from these three datasets, t-SNE \cite{van2008visualizing} is employed to visualize the distribution of the datasets in Fig. \ref{fig:t-SNE}, where domain gap between tumors of cross-organs can be observed.
	
	\subsection{Experimental Settings}
	In this experimental setup, the learning strategy is utilizing a learning rate of 10e-4, weight decay of 0.05. $\alpha$ and $\beta$ are set to $0.5$. The weight value $\lambda$ is set 0.9. The AdamW \cite{loshchilov2017decoupled} optimizer is employed for optimization purposes. The batch size for each iteration of gradient update is set to 4 and the input size for the model is specified as $256 \times 256$  pixels. We train 200 epochs for all methods inside our COTS-Nets.
	
	We utilize U-shaped ViT based on the architecture of U-Net with a domain adapter as our universal network, same as MDViT \cite{du2023mdvit, lee2022mpvit}.
	For a fair comparison in the training process, 401 images are randomly selected from Thyroid\_tn3k or MMOTU\_2d as the source domain, and 401 images from the PCEUS dataset are used as the target domain. These images are used for adaptation experiments in both the “mmotu $\rightarrow$ pancreatic” and “thyroid $\rightarrow$ pancreatic” directions, which means ``source $\rightarrow $ target''.
	To prevent overfitting and improve the robustness of the model, data augmentation techniques are used to increase the variability of the training data. Examples include Gaussian noise, flipping the image, randomly adjusting brightness and contrast, randomly panning, scaling, and rotating the image.
	During testing process, the remaining 100 images of the PCEUS dataset are used to verify the performance of the model.
	
	All the computations are performed on a GPU server with NVIDIA GeForce GTX 3090, Intel Core i7-13700F CPU processor.
	\begin{figure*}[!t]
		\centering
		\includegraphics[width=17.3cm]{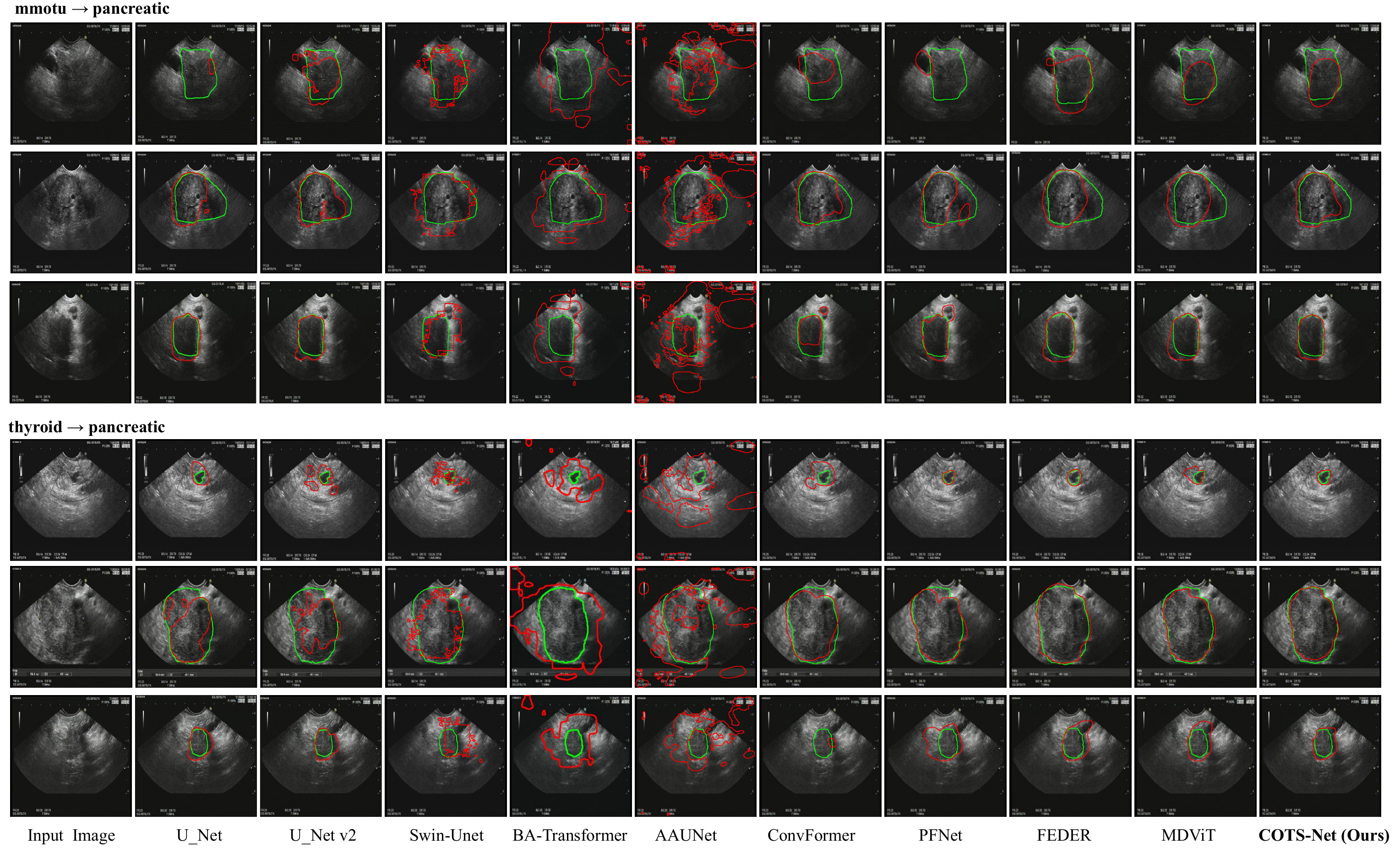}
		\caption{Segmentation results of different methods on ``mmotu $\rightarrow$ pancreatic'' and ``thyroid $\rightarrow$ pancreatic''. From top to bottom, the top two lines are ``mmotu $\rightarrow$ pancreatic'' and the bottom two lines are ``thyroid $\rightarrow$ pancreatic''. From left to right are images of input, U\_Net, U\_Net v2, Swin\_Unet, BA-Transformer, AAUNet, ConvFormer, PFNet, FEDER, MDViT and COTS-Nets (Ours). The green contours represent the ground-truth labels, while the red contours represent the predicted segmentation. The images are EUS of pancreatic lesions, showing the network's performance in accurately segmenting tumors using different cross-organ cases. }
		\label{fig:Results_Visual}
	\end{figure*}

	\subsection{Quantitative Evaluation Metrics}
	
	For a fair comparison, we employ four widely adopted segmentation metrics to quantitatively evaluate the performance of different methods in cross-organ medical image segmentation. These metrics include Dice coefficient, Intersection over Union (IoU), Average Symmetric Distance (ASD) and 95 $\%$ Hausdorff Distance (HD95) \cite{feng2023unsupervised, vesal2022domain}.
	
	% 修改
	\textbf{Dice coefficient:} The Dice coefficient measures the overlap between the predicted segmentation and the ground-truth, providing a value between 0 and 1. A higher Dice value indicates a greater overlap and, consequently, better segmentation performance.
	
	\textbf{Intersection over Union (IoU):} IoU, also known as the jaccard index, quantifies the ratio of the intersection to the union of the predicted segmentation and the ground-truth. Similar to Dice, higher IoU values indicate better segmentation accuracy.

	\textbf{Average Symmetric Distance (ASD):} ASD calculates the average of the shortest distances between the boundary points of the predicted segmentation and the ground-truth, measured symmetrically. Lower ASD values suggest that the predicted boundary closely aligns with the ground-truth boundary, indicating superior segmentation quality.
	
	\textbf{95 $\%$ Hausdorff Distance (HD95):} In order to exclude unreasonable distances caused by some outliers and to maintain overall numerical stability, the top 95$\%$ of the distances ranked from smallest to largest were selected as the actual Hausdorff distances and called HD95. A lower HD95 value signifies that the predicted boundary closely matches the ground-truth boundary, reflecting better performance.
	
	In summary, higher values of Dice and IoU indicate better segmentation results, demonstrating greater overlap and accuracy. 
	Conversely, lower values of ASD and HD95 represent superior performance, reflecting more precise boundary alignment between the predicted segmentation and the ground-truth. These metrics together provide a comprehensive assessment of segmentation quality.

	\subsection{Comparison Study}
	To verify the effectiveness of COTS-Nets, we compare our method with several state-of-the-art methods including U-Net \cite{ronneberger2015u}, U-Net V2 \cite{peng2023u}, Swin-Unet \cite{hatamizadeh2021swin}, BA-Transformer \cite{wang2021boundary}, ConvFormer \cite{lin2023convformer}, all of which are typical medical image segmentation methods. Meanwhile, AAUNet and MDViT \cite{chen2022aau, du2023mdvit} combined with domain adaptation is also compared. 
	Considering the low contrast of the recognized targets in the ultrasound images, which is similar to some extent to the camouflaged target detection, we also choose FEDER and PFNet camouflaged target segmentation algorithms as comparison algorithms \cite{mei2021camouflaged, he2023camouflaged}.
	
	\subsubsection{Compare with the SOTA methods} From the results in Table \ref{Table1}, we can observe that the performance of the proposed COTS-Nets is superior to the compared SOTA methods. 
	For instance, under the “mmotu $\rightarrow$ pancreatic” setting, our method improves by 1.094\% on Dice and 1.592\% on IOU, respectively, compared to MDViT, while it has 1.790\% and 2.469\% performance improvements under the “thyroid $\rightarrow$ pancreatic” setting, which means the COTS-Nets learns properly homogenization information from the cross-organ to guide the process of segmenting PCEUS dataset.
	
	Although the camouflage target detection algorithms PFNet and FEDER still maintain a high generalization performance in the medical image-oriented space and are able to effectively identify tumors in pancreatic endoscopic ultrasound images, the ASD and HD\_95 metrics are far inferior to those of our proposed COTS-Nets algorithm, COTS-Nets is able to segment the boundaries of target tumors more accurately.
	
	Meanwhile, under the ``mmotu $\rightarrow$ pancreatic'' direction, the ASD of our method is improved by 0.20mm compared to MDViT, indicating that COTS-Nets can delineate pancreatic tumor boundaries more accurately with a smaller average surface distance from the ground-truth. Specifically, the ASD of COTS-Nets is 0.42mm, which is 0.20mm lower than the 0.62mm of MDViT. Meanwhile, the HD95 of COTS-Nets is 4.76mm. In the ``thyroid $\rightarrow$ pancreatic'' direction, our method improves the ASD value by 0.12mm over MDViT. The ASD value of COTS-Nets is 0.71mm, which is 0.12mm lower than that of 0.83mm recorded by MDViT. Also, the HD95 of COTS-Nets is 1.99mm lower than MDViT's. This further demonstrates that COTS-Nets can effectively utilize the information of cross-organ tumors to improve segmentation accuracy and provide more precise and reliable and provide a more accurate and reliable predictions of pancreatic tumor boundaries.

	\subsubsection{Visualization of segmentation results} Fig. \ref{fig:Results_Visual} shows some qualitative visual segmentation examples on direction of ``mmotu $\rightarrow$ pancreatic'' and ``thyroid $\rightarrow$ pancreatic''. 
	It is observed that: 1) Compared with other methods, COTS-Nets produces more accurate results in both ``mmotu → pancreatic'' and ``thyroid → pancreatic'' tasks, effectively capturing the tumor boundaries with minimal errors; 2) Our approach consistently delineates clear boundaries and closely matches the ground-truth, outperforming the other methods in both visual accuracy and consistency.
	
	Compared with the segmentation results of other methods, our method not only effectively mitigates the perturbations of tumor size, surrounding tissues, and cascades, but also obtains segmentation results that are closer to the ground-truth mask. In addition, the method proposed in this paper also alleviates the effect of heterogeneous structures on segmentation results and conversely enhances the lesion segmentation of pancreatic tumors by exploiting the homogeneous features of cross-organ tumors. The comprehensive evaluation results and visualization show that our method can achieve the best segmentation results in pancreatic lesion segmentation with less leakage and misdetection.

	\subsection{Ablation}
	\subsubsection{Significance of cross-organ dataset} 
\begin{figure}[!t]
		\centering
		\begin{subfigure}{1.0\linewidth}
			\centering
			\includegraphics[width=0.72\linewidth]{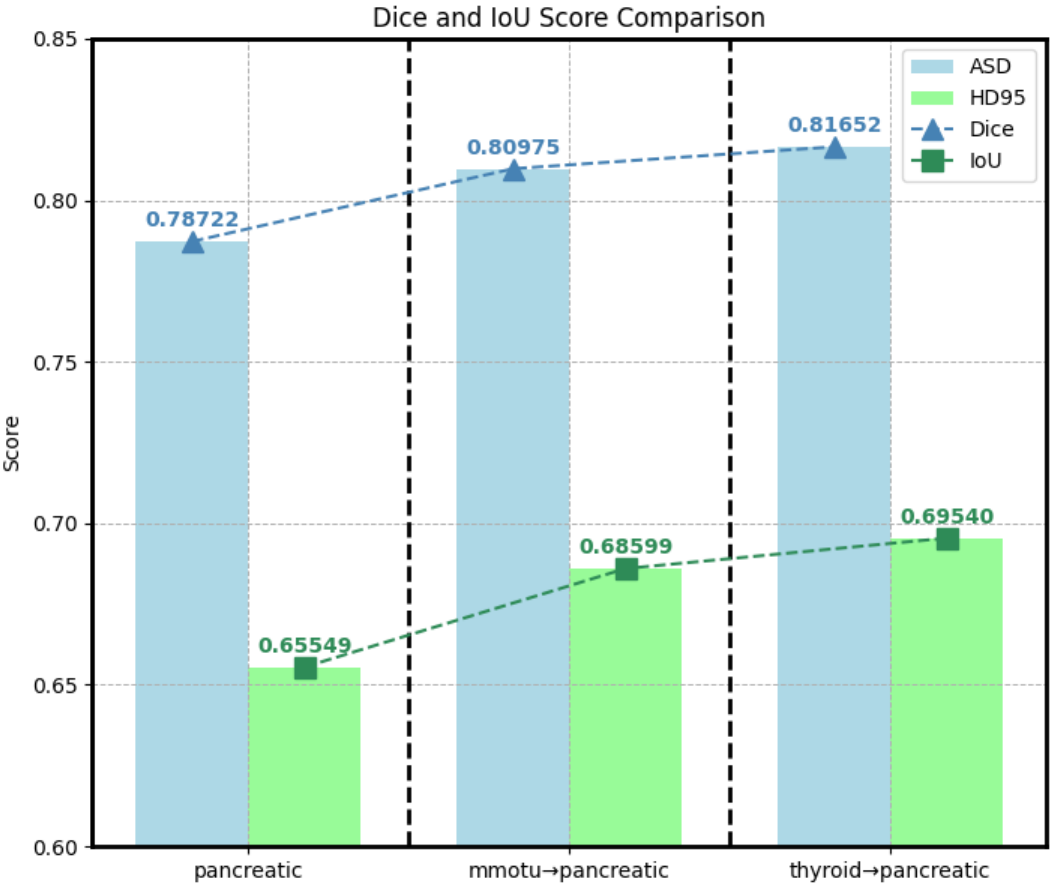}
			\captionsetup{font=scriptsize}
			\caption{The metric of Dice and IoU}
			\label{fig:Pancreatic_Ablation_1}%文中引用该图片代号
		\end{subfigure}
		\centering
		\begin{subfigure}{1.01\linewidth}
			\centering
			\includegraphics[width=0.69\linewidth]{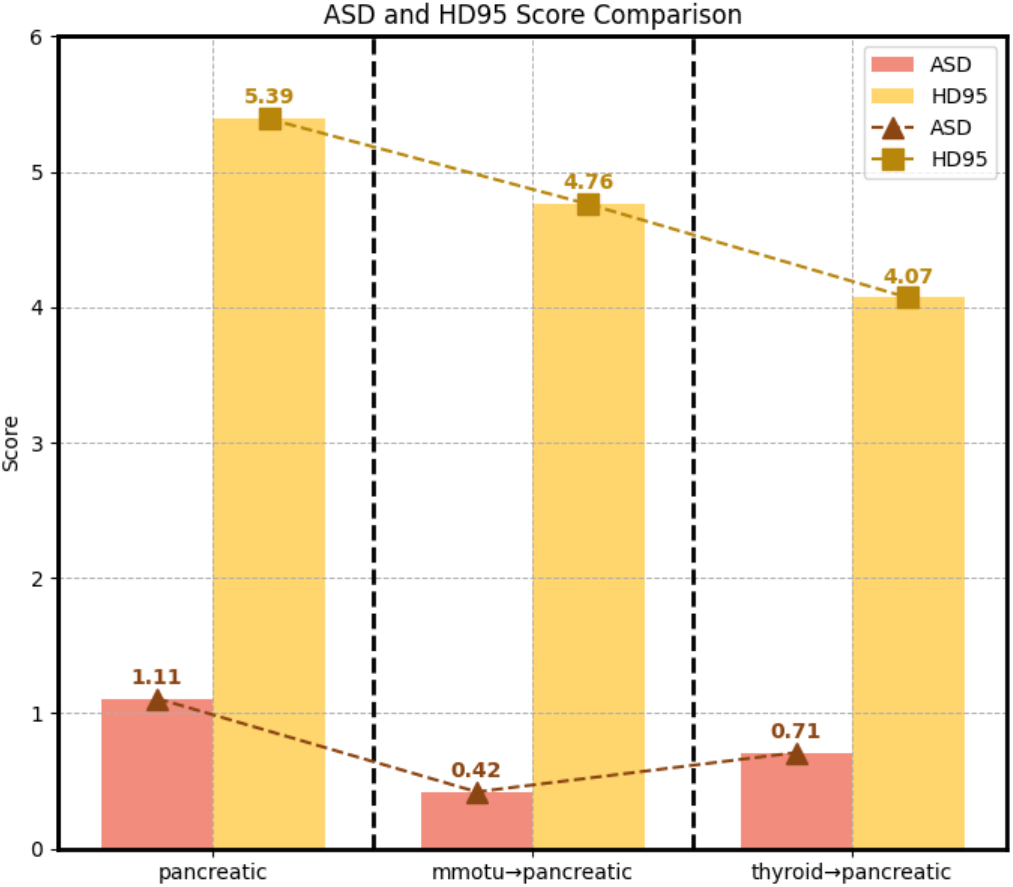}
			\captionsetup{font=scriptsize}
			\caption{The metric of ASD and HD95}
			\label{fig:Pancreatic_Ablation_2}%文中引用该图片代号
		\end{subfigure}
		\caption{Ablation results of single pancreatic endoscopic ultrasound and cross-organ pancreatic endoscopic ultrasound tumor detection.}
		\label{fig:Pancreatic_Ablation}
	\end{figure}

To validate that the addition of extra datasets as auxiliary feature supplements can be decisively instrumental in pancreatic endoscopy ultrasound images for tumor segmentation, we examine the impact of cross-organ in COTS-Nets. We compare the case of cross-organ with single PCEUS dataset as Fig. \ref{fig:Pancreatic_Ablation} shows. To be brief, we investigate the effects of different cross-organ combinations, denoted as ``pancreatic'', ``mmotu $\rightarrow$ pancreatic'' and ``thyroid $\rightarrow$ pancreatic'' to represent different cross-organ combinations respectively. Additionally, we analyze the impact of varying weights for cross-organ datasets on pancreatic tumor segmentation. Similarly, the directions denote as ``200\_mmotu $\rightarrow$ pancreatic'' and ``200\_thyroid $\rightarrow$ pancreatic'', which are 200 randomly selected images from MMOTU\_2d and Thyroid\_tn3k. Note that all individual cases are trained using the same parameters as COTS-Nets to ensure a fair comparison.
	
	As shown in Fig. \ref{fig:Pancreatic_Ablation_1}, we use Dice and IoU as overlap-based metrics. Our COTS-Nets achieves a Dice score of 78.721\% and an IoU of 65.549\% for pancreatic ultrasound endoscopic lesion segmentation using only a single PCEUS dataset. In the case of ``mmotu $\rightarrow$ pancreatic'' and ``thyroid $\rightarrow$ pancreatic'', Dice of 80.975\% and 81.625\%, along with IoU scores of 68.599\% and 69.504\%. These results are significantly higher than those obtained when PCEUS is used as a single data input. This experiment effectively demonstrates that tumors from different organs share common homogeneous features, which can be leveraged as complementary information to improve the performance of lesion segmentation, particularly in scenarios with low contrast and high noise.
	
	Additionally, we use ASD and HD95 as distance-based metrics, as shown in Fig. \ref{fig:Pancreatic_Ablation_2}. Using only a single PCEUS dataset, we obtain an ASD of 1.11mm and an HD95 of 5.39mm, indicating a significant distance margin between the predicted boundary and the ground-truth boundary in COTS-Nets. Similarly, for ``thyroid $\rightarrow$ pancreatic'', ASD is 0.71mm and HD95 is 4.07mm. These results indicate a more accurate predictive boundary compared to using only the PCEUS dataset. These distance-based metrics validate that cross-organ ancillary data can effectively address the issue of unclear boundaries in pancreatic endoscopic ultrasound images, thereby improving the accuracy of tumor boundary prediction.
	
	\begin{table}[!t]
		\centering
		\renewcommand\arraystretch{1.3}
		\caption{Ablation study on cross-organ datasets with different weights. The direction with ``200'' implies that 200 cross-organ datasets are randomly selected.}
		\label{Table2}
		\resizebox{8.6cm}{!}
		{
			\begin{tabular}{c|cccc}
				\hline
				\textbf{Datasets} & \textbf{Dice} & \textbf{IoU} & \textbf{ASD} & \textbf{HD95}\\
				\hline
				pancreatic & 78.722 & 65.549 & 1.11 & 5.39\\
				\hline
				200\_mmotu $\rightarrow$ pancreatic & 79.403 & 66.433 & 0.77 & 6.15\\
				mmotu $\rightarrow$ pancreatic & \textbf{80.975} & \textbf{68.599} & \textbf{0.42} & \textbf{4.76} \\
				\hline
				200\_thyroid $\rightarrow$ pancreatic & 80.209 & 67.588 & 0.88 & 4.95\\
				thyroid $\rightarrow$ pancreatic & \textbf{81.652} & \textbf{69.504} & \textbf{0.71} & \textbf{4.07}\\
				\hline
			\end{tabular}
		}
	\end{table}
	
	The results presented in Table. \ref{Table2} demonstrate that the recognition accuracy of COTS-Nets improves as the proportion of cross-organ datasets increases, further indicating that incorporating cross-organ tumors effectively enhances the features of pancreatic tumors. However, the inclusion of cross-organ tumor data introduces a bias in the model, as evidenced by an increase in HD95 when 200 ovarian images are added. Nonetheless, by balancing the data size, COTS-Nets effectively manages the domain gap between different organs and enhances the segmentation accuracy of pancreatic tumors.
	\begin{table}[!t]
		\centering
		\renewcommand\arraystretch{1.25}
		\caption{Ablation study on ``thyroid$\rightarrow$pancreatic''}
		\label{Table3}
		\resizebox{9cm}{!}
		{
			\begin{tabular}{c|cccc|cc}
				\hline
				\# & Baseline & HAAM & $\mathcal{L}_{consist}$ & $\mathcal{L}_{bound}$ & \textbf{Dice} & \textbf{IoU} \\
				\hline
				1 & \Checkmark & & & & 79.862 & 67.035 \\
				2 & \Checkmark & \Checkmark & & & 80.546 & 67.941 \\
				3 & \Checkmark & \Checkmark & \Checkmark & & 80.601 & 67.934 \\
				4 & \Checkmark & \Checkmark & \Checkmark & \Checkmark & \textbf{81.652} & \textbf{69.504} \\
				\hline
			\end{tabular}
		}
	\end{table}
	
	\subsubsection{Importance of different modules.} To further evaluate the scalability of our model, we explore the importance and impact of different modules in COTS-Nets. We have designed four separate cases in the direction  “thyroid $\rightarrow$ pancreatic” to examine the performance when applying different combinations. Note that all separate cases are trained using the same parameters as COTS-Nets for fair comparison.

	\begin{itemize}
		\item[$\bullet$] Baseline: Only using $\mathcal{L}_{seg}$ and $\mathcal{L}_{distill}$ for PCEUS segmentation .
		\item[$\bullet$] Baseline + HAAM: Extract channel and spatial represent features of the specific domain for universal network training with HAAM module.
		\item[$\bullet$] Baseline + $\mathcal{L}_{consist}$ + HAAM: Execute a measure of the consistency of the model's output with cross-organ tumor data using $\mathcal{L}_{consist}$.
		%模型在不同输入扰动下输出的一致性的度量
		\item[$\bullet$] Baseline + $\mathcal{L}_{consist}$ + $\mathcal{L}_{bound}$ + HAAM (final). Using $\mathcal{L}_{bound}$ to focus on boundary region differences between cross-organ tumor segmentation, with a concentration on optimizing the boundary between predicted segmentation results and true labels.
	\end{itemize}

%	\begin{table}[!h]
%		\centering
%		\renewcommand\arraystretch{1.4}
%		\caption{Ablation study on ``thyroid$\rightarrow$pancreatic''}
%		\label{Table2}
%		\resizebox{9cm}{!}
%		{
%			\begin{tabular}{c|l|c|cc}
%				\hline
%				\# & \textbf{Methods} & \textbf{Dice} & \textbf{IoU} \\
%				\hline
%				1 & Baseline & 79.862 & 67.035 \\
%				2 & Baseline + HAAM & 80.546 & 67.941 \\
%				3 & Baseline + $\mathcal{L}_{consist}$ + HAAM & 80.601 & 67.934 \\
%				4 & Baseline + $\mathcal{L}_{consist}$ + $\mathcal{L}_{boundary}$ + HAAM & \textbf{81.652} & \textbf{69.504} \\
%				\hline
%			\end{tabular}
%		}
%	\end{table}

	The results are reported in Table \ref{Table3}. Overall, our proposed COTS-Nets brings the most significant performance with 1.790\% Dice and 2.469\% IoU improvement over the Baseline.
	
	\textbf{Significance of HAAM}: We first apply the HAAM module to extract specific features at different scales from cross-organ. HAAM brings a beneficial effect — Dice and IoU improve 0.648\% and 0.906\% respectively (\#1vs.\#2). 
	The results confirm that HAAM in auxiliary networks is able to learn generalized representational features, which is crucial for the universal network learning.
	However, the more pronounced improvement in IoU compared to Dice indicated that HAAM may be more about optimizing the overall segmentation region rather than the details and boundaries.
	
	\textbf{Importance of $\mathcal{L}_{consist}$}: Considering that different medical images across organs may introduce disturbing factors such as scanning angles, lighting conditions, etc. $\mathcal{L}_{consist}$ is thus used to maintain the generalization performance of the images in universal network. By comparing \#2vs.\#3, we observe that the Dice is improved by 0.055\% and IoU is decreased by 0.007\%.
	This result suggests that the introduction of $\mathcal{L}_{consist}$ is a model that performs better on the overall region.
	Hence, we should focus more on the details of the lesion boundary region to achieve more accurate PCEUS lesion segmentation.

	\textbf{Effective of $\mathcal{L}_{bound}$}: $\mathcal{L}_{bound}$ is used in the universal network of COTS-Nets to enhance the accuracy of the segmentation boundary. This loss function specifically focuses on optimizing the precision of the segmentation at the boundary regions of the lesions. $\mathcal{L}_{bound}$ learns the homogeneity of cross-organ tumors using the respective boundary maps, ensuring that the model pays special attention to the edges and contours of the lesions. $\mathcal{L}_{bound}$ shows a significant improvement in both Dice and IoU metrics (\#3vs.\#4), with 1.051\% and 1.570\% improvement respectively. By integrating $\mathcal{L}_{bound}$, COTS-Nets meticulously captures the homogenized boundary details required for accurate lesion segmentation in cross-organ datasets with PCEUS images.
	
	\section{Conclusion}
	In our work, COTS-Nets exploits consistency loss and boundary loss within the universal network. This approach enhances the segmentation accuracy of PCEUS by leveraging  boundary information in additional organ tumors from both overall and local perspectives.
	Notably, the introduction of consistency loss also helps mitigate the effects of the domain gap.
%	To further alleviate cross-organ gap, the auxiliary network guides the universal network learning so as to distill the domain-invariant knowledge to alleviate the domain gap. It enables COTS-Nets to effectively learn the homogeneity of tumors across different organs, which significantly improves the segmentation performance of the model for pancreatic tumors. 
	To further alleviate cross-organ gap, the auxiliary network enhances representation learning by fusing multi-scale features so as to refine domain-invariant knowledge in the auxiliary network. It enables COTS-Nets to effectively learn the homogeneity of tumors across different organs, which significantly improves the segmentation performance of the model for pancreatic tumors. 
%	Moreover, consistency loss and boundary loss enhance the segmentation accuracy of PCEUS by leveraging additional organ tumors from both overall and local perspectives.
	Extensive experiments demonstrate that COTS-Nets can optimize the representation of tumors and the proposed method can outperform existing medical image segmentation methods, supporting the notion that tumors across different organs have homogeneity in the medical field.
	
%	\clearpage

\end{document}